\documentclass[conference]{IEEEtran}
\IEEEoverridecommandlockouts
\usepackage{cite}
\usepackage{amsmath,amssymb,amsfonts}
\usepackage{algorithmic}
\usepackage{graphicx}
\usepackage{textcomp}
 \usepackage{xcolor}

\usepackage{algorithm}
\usepackage{algorithmic}
\usepackage{bm}
\usepackage{subfigure}
\def\BibTeX{{\rm B\kern-.05em{\sc i\kern-.025em b}\kern-.08em
    T\kern-.1667em\lower.7ex\hbox{E}\kern-.125emX}}
\begin{document}

\title{A fast online cascaded regression algorithm for face alignment\\
% {\footnotesize \textsuperscript{*}Note: Sub-titles are not captured in Xplore and
% should not be used}
\thanks{This study was funded by National Natural Science Foundation of People's Republic of China (No.61672130, No.61602082, No.91648205), the National Key Scientific Instrument and Equipment Development Project (No. 61627808), the Development of Science and Technology of Guangdong Province Special Fund Project Grants (No. 2016B090910001), the LiaoNing Revitalization Talents Program (No. XLYC1806006).}
}

% \author{\IEEEauthorblockN{1\textsuperscript{st} Lin Feng}
% \IEEEauthorblockA{\textit{School of Innovation and Entrepreneurship} \\
% \textit{Dalian University of Techonology}\\
% Dalian, China \\
% e-mail: fenglin@dlut.edu.cn}
% \and
% \IEEEauthorblockN{2\textsuperscript{nd} Caifeng Liu}
% \IEEEauthorblockA{\textit{Faculty of Electronic Information and Electrical Engineering} \\
% \textit{Dalian University of Techonology}\\
% Dalian, China  \\
% liucaifeng12345@mail.dlut.edu.cn}
% \and
% \IEEEauthorblockN{3\textsuperscript{rd} Shenglan Liu}
% \IEEEauthorblockA{\textit{School of Innovation and Entrepreneurship} \\
% \textit{Dalian University of Techonology}\\
% Dalian, China \\
% liusl@mail.dlut.edu.cn}
% \and
% \IEEEauthorblockN{4\textsuperscript{th} Huibing Wang}
% \IEEEauthorblockA{\textit{College of Information Science and Technology} \\
% \textit{Dalian Maritime University}\\
% Dalian, China \\
% whb08421005@mail.dlut.edu.cn}

% \and

\author{\IEEEauthorblockN{Lin Feng\IEEEauthorrefmark{1}, Caifeng Liu\IEEEauthorrefmark{2}, Shenglan Liu\IEEEauthorrefmark{1}, Huibing Wang\IEEEauthorrefmark{3}}
\IEEEauthorblockA{\IEEEauthorrefmark{1}School of Innovation and Entrepreneurship\\
Dalian University of Techonology, Dalian, China\\
Email: \{fenglin, liusl\}@dlut.edu.cn}
\IEEEauthorblockA{\IEEEauthorrefmark{2}Faculty of Electronic Information and Electrical Engineering\\
Dalian University of Techonology, Dalian, China\\
liucaifeng12345@mail.dlut.edu.cn}

\IEEEauthorblockA{\IEEEauthorrefmark{3}College of Information Science and Technology\\
Dalian Maritime University, Dalian, China\\
whb08421005@mail.dlut.edu.cn}}

\maketitle

\begin{abstract}
Traditional face alignment based on machine learning usually tracks the localizations of facial landmarks employing a static model trained offline where all of the training data is available in advance.
When new training samples arrive, the static model must be retrained from scratch, which is excessively time-consuming and memory-consuming.
In many real-time applications, the training data is obtained one by one or batch by batch. It results in that the static model limits its performance on sequential images with extensive variations. 
Therefore, the most critical and challenging aspect in this field is dynamically updating the tracker's models to enhance predictive and generalization capabilities continuously. In order to address this question, we develop a fast and accurate online learning algorithm for face alignment. Particularly, we incorporate on-line sequential extreme learning machine into a parallel cascaded regression framework, coined incremental cascade regression(ICR). To the best of our knowledge, this is the first incremental cascaded framework with the non-linear regressor. One main advantage of ICR is that the tracker model can be fast updated in an incremental way without the entire retraining process when a new input is incoming. Experimental results demonstrate that the proposed ICR is more accurate and efficient on still or sequential images compared with the recent state-of-the-art cascade approaches. Furthermore, the incremental learning proposed in this paper can update the trained model in real time.
\end{abstract}


\section{Introduction}
\label{sec:Introduction}

Face alignment aims to locate a sparse set of facial landmarks for a given facial image or video. It is a topic of interest in the domain of Computer Vision because many subsequent face analysis tasks, such as face recognition \cite{wu2019cross}, facial animation, and authentication on the Internet of Things \cite{qiu2018eabs}, heavily depend on the accurate localizations of facial landmarks. Over the decades, various face alignment procedures have been proposed, which can be broadly classified into generative models and discriminative models. The generative approaches adopt an analysis-by-synthesis loop where the optimization strategy attempt to find the optimal shape parameters by maximizing the joint posterior probability between the pre-built deformable model and feature of the input image. The representative examples of this category are Active Appearance Model(AAM) \cite{cootes2001active} and Gauss-Newton Deformable Part Model(GN-DPM) \cite{tzimiropoulos2014gauss}.

The discriminative models seek to learn discriminative information (i.e. discriminative function \cite{liu2009discriminative,xiong2013supervised}) which directly maps representation of facial appearance to facial landmarks. Many discriminative methods utilize popular Cascaded Regression(CR) framework, in which a series of regressors are learned in a cascaded manner to gradually refine the initialization to ground-truths. Numerous cascaded regression methodologies have shown to produce excellent results on face alignment tasks and validate the CR framework's superior efficiency and accuracy \cite{shi2018face,lee2015face,asthana2014incremental,tzimiropoulos2015project,cui2018recurrent}. The most efficient of these methods is LBF \cite{ren2014face} using a set of local binary features to learn a cascade linear regressors and
its running speed can be achieved over 3000fps on a standard desktop for locating a few dozens of landmarks. The authors of \cite{xiong2013supervised} attempt to provide a theoretical explanation of cascade linear regression from the perspective of least squares optimization and solve it as a supervised descent methodology. While these cascaded linear regression methods are very efficient, they are suffering in poor fitting capability to exploit the non-linear and complex relationship between feature space and shape variations in unconstrained scenarios.

Considering the limitations of linear regression, some non-linear regressors based on decision-making tree, such as boosting \cite{cao2014face} and random forest \cite{dollar2010cascaded} were introduced into cascaded regression. However, these ensemble learning models are prone to over-fitting and suffered from very high computational burden \cite{jin2017face}.
With the research wave of deep learning \cite{wu2019cycle} has been carved out in the image domain, many deep-based methods \cite{shi2018face,jourabloo2017pose,wang2017effective,wu20193} have been proposed and achieved breakthrough successes in some big scale datasets. Because of their complicated structure and a great number of hyperparameters, these deep frameworks tend to consume massive time to train the models. Moreover, deep structure encounters a complete retraining process if the training data is supplemented.

Although these discriminative methods have accomplished superior performance for face alignment under unconstrained faces or some more challenging situations, they are limited by a static generic model that is built completely on offline training data. Nevertheless, such a static model can not be updated in real time to handle some certain specific tasks(e.g. person-specific landmarks tracking for video). 
Since the entire training procedure is very time-consuming and very expensive, how to best exploit discriminative cascade regression for incremental learning is an intractable issue.
A few studies \cite{asthana2014incremental,sanchez2018functional} have started to work on this vital issue from the viewpoint of the incremental linear regression function. However, linear regressor tends to be limited to a linear relationship between dependent and independent variables.

In order to overcome the aforementioned limitations, in this paper, we study incremental training of cascaded regression with non-linear regressor.
As the foundation of our algorithm, CR framework is one of the most practical and effective framework for localizing facial landmarks. Nevertheless,
traditional CR framework still has two limitations to achieve incremental training. (1) It successively trains a series of regressors stage by stage. The entire procedure(4 or more cascaded stages) is too slow to satisfy the requirements of online learning in real time. (2) In the cascade executions, the input of each stage intensely depends on the outputting shapes of the previous stage. In that case, if the certainly stage-regressor is incrementally updated, the whole input set of the subsequent stage will be recomputed by new regressor and all samples formerly trained must be reloaded. Obviously, these limitations can lead to a vast resource-consuming and time-consuming when the scale of data is so large.

\begin{figure}
    \centering
    \begin{tabular}{c}
        \includegraphics[height=4.5cm]{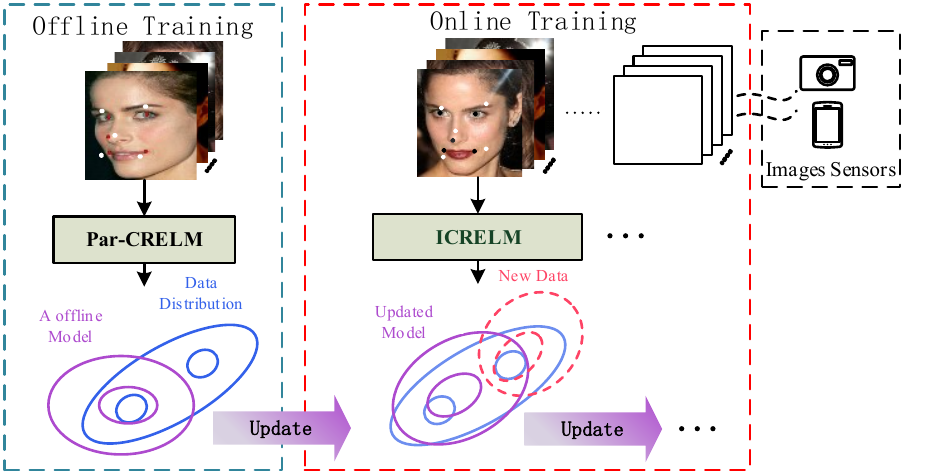}
    \end{tabular}
    \caption{Overview of our approach}
    \label{fig:Whole framework of proposed method}
\end{figure}

For these, we propose incremental cascade regression(ICR), which aims to train and update the series of non-linear regressors in a parallel manner instead of a sequential one. In special, we adopt Monte Carlo sampling methodology \cite{xiong2013supervised,asthana2014incremental} to approximate the shape space, in which the facial shape is no longer depend on the outputting of the previous stage. Meanwhile, ICR is equipped extreme learning machine(ELM) as the discriminative regressor to learn the mapping between facial feature representations and the shape variations.
ELM has powerful capability to approximate any linear or non-linear mapping(e.g. the least square constraints in face alignment).
Moreover, ELM has very fast training speed and low computational cost without the hassle of parameters tuning when compared to the gradient descent-based regressors or decision-tree based regressors. As shown in Figure~\ref{fig:Whole framework of proposed method},
ICR divides into two parts: offline and online training procedures.
In the offline training procedure, a generic model can be learned by a parallel cascade regression of ELM.
Then the online training procedure can update the trained model by using the Monte Carlo sampling methodology \cite{xiong2013supervised,asthana2014incremental}. In this way, incrementally updating the trained regressor of each stage does not depend on the outputting of the previous stage, so we can update all the regressors in parallel. 
Meanwhile, we adopt an online sequential extreme learning machine(OS-ELM) \cite{liang2006fast} method to update the trained ELM regressors. 
The OS-ELM \cite{liang2006fast} is an incremental learning strategy of extreme learning machine(ELM) \cite{huang2012extreme}.
In summary, our main contributions are as follows:
\begin{itemize}
\item To the best of our knowledge, ICR is the first parallel cascaded regression framework equipped the non-linear regressors.
\item ICR is capable of replenishing new training data and very fast updating the model without retraining from scratch, which can constantly increase the generalization and robustness of model.
\item We evaluate ICR on three datasets and demonstrate the importance of incremental learning in achieving state-of-the-art performance on sequential training data.
\end{itemize}

\section{Related work}
\label{sec:Related works}

As described above, the existing methods can be split into generative and discriminative categories. Both categories have proposed diverse models for offline face alignment with varying degrees of success. The main problem that our method can address is incremental learning for face alignment. In this section, we will focus on closely related works on this task.  

On the generative side, very limited research has been introduced to AAM \cite{sung2009adaptive}, in which the incremental principal component analysis(iPCA) \cite{levey2000sequential} is used to update the generic AAM's linear appearance model with current face image. However, this method heavily relys on prebuilding a robust parametric model as well as AAM and need images of the same person for training, which is less generalization and unpractical.

On the discriminative side, Asthana \emph{et al.} \cite{asthana2014incremental} proposed a incremental version of SDM called iPar. In this work, a parallel strategy is used to implement the incremental update of the modified linear regressors. For each cascade stage, they utilize a set of perturbations from a pre-populated Gaussian distribution instead of the outputting shape variations of the previous stage. In this way, computing each regressor is independent of the other stages, which can facilitate the whole training procedure in a parallel manner. As experimentally shown by \cite{asthana2014incremental}, such approximate training strategy achieves similar accuracy as the model of SDM trained in a sequential manner. This study offers a new idea and premise for fast training incrementally discriminative regressors. However, 
% on the one hand, iPar ignores the updating of the statistical distribution when new training data is added, which may result in that the sampling space is limited in the distribution generated by the initial training set. On the other hand, the 
linear regressors used in iPar are too weak to exploit the complex relationship between shape variables and appearance features.

Inspired by \cite{asthana2014incremental}, we propose a incremental cascade regression, coined ICR, that pay more attention to non-linear discriminative regression in a parallel cascaded regression framework.

\section{Method}
\label{sec:Method}

In this section, we present a cascade regression of extreme learning machine and its online training version. As a preliminary, we first take a brief review the ELM algorithm in detail. 
\subsection{Extreme Learning Machine}
ELM is an efficient way of building the single layer feed forward neural networks(SLFNs) \cite{zhang2018robust}. 
Given $N$ input-output training samples, arbitrary distinct samples $\bm{(}{\bm{x}_i},{\bm{y}_i}) \in {\Re ^n} \times {\Re ^m}$. Here, $\bm{x}_i$ is a $n \times 1$ input vector and $\bm{y}_i$ is a $m \times 1$ target vector. ELM with $K$ hidden nodes and activation $G( \cdot )$ can be mathematically modeled as 
\begin{equation}
\label{eq:ELM_K_equations}
     \sum\limits_{k = 1}^K {{\bm \beta _k}G(\bm{a}{}_k,{b_k},{\bm{x}_i})}  = {\bm{y}_i},{\rm{   i = 1,2,}} \ldots {N},
\end{equation}
where ${{\bm{a}}_k} \in {\Re ^n}$ and ${b_i} \in \Re $ are the randomly chosen learning parameters of hidden nodes, $G({\bm{a}_k},{b_k},\bm{x})$ is the output of $i$-$th$ hidden nodes w.r.t the input $\bm{x}$ for additive units with the activation function $g( \cdot )$(e.g., sigmoid), $G(\cdot)$ is defined as
\begin{equation}
  G({\bm{a}_k},{b_k},\bm{x}) = g(\bm{a_{k}} \cdot \bm{x}+b_{k}),  
\end{equation}
where the $\bm{a_{k}} \cdot \bm{x}$ denotes the inner product of vectors $\bm{a_{k}}$ and $\bm{x}$ in ${\Re ^n}$. The compact form of $K$ equations in Equation(\ref{eq:ELM_K_equations}) is: 
\begin{equation}
\label{eq:ELM_matrix_representation}
    \mathbf{H} \mathbf{ \beta}  = \mathbf{Y},
\end{equation}
where 
\begin{equation}
\label{eq:representation of H}
    \begin{array}{l}
{\bf{H}}({\bm{a}_1}, \ldots ,{\bm{a}_K},{b_1}, \ldots ,{b_K},{\bm{x}_1}, \ldots ,{\bm{x}_N})\\
{\rm{  }} = {\left[ \begin{array}{l}
G({\bm{a}_1},{b_1},{\bm{x}_1}) \ldots G({\bm{a}_K},{b_K},{\bm{x}_1})\\
{\rm{         }} \vdots {\rm{       }} \cdots {\rm{          }} \vdots \\
G({\bm{a}_1},{b_1},{\bm{x}_N}) \ldots G({\bm{a}_K},{b_K},{\bm{x}_N})
\end{array} \right]_{N \times K}}
\end{array}
\end{equation}
\begin{equation}
    {\bf{\beta }}\bm{ = }{\left[ \begin{array}{l}
{\bm{\beta }_1}^T\\
{\rm{ }} \vdots \\
{\bm{\beta }_K}^T
\end{array} \right]_{K \times m}}, {\bf{Y}}{\rm{ = }}{\left[ \begin{array}{l}
{\bm{y}_1}^T\\
{\rm{ }} \vdots \\
{\bm{y}_N}^T
\end{array} \right]_{N \times m}}.
\end{equation}

${\bf{H}}$ is the out matrix of hidden layer, where the $k$-$th$ column of ${\bf{H}}$ is the output vector of $k$-$th$ hidden node with respect to inputs ${\bm{x}_1},{\bm{x}_2}, \ldots ,{\bm{x}_N}$. In the Equation(\ref{eq:ELM_matrix_representation}), the learning parameters of hidden layer nodes can be randomly generated, So the output weights $\bf{\beta}$ can be estimated by finding the least square solution of the linear system, according to the expression:
\begin{equation}
\label{eq:ELM solution_ordinary}
    {\bf{\tilde \beta }} = \mathop {\arg \min }\limits_{\bf{\beta }} ||{\bf{H\beta }} - {\bf{Y}}||_2^2 = {{\bf{H}}^\dag }{\bf{Y}},
\end{equation}
where ${{\bf{H}}^\dag }$ is the Moore-Penrose gneralized inverse of $\bf{H}$. In practical, it is usually comes that $N>K$. The Equation(\ref{eq:ELM solution_ordinary}) can be rewritten as 
\begin{equation}
\label{eq:ELM solution nonsingular}
    {\bf{\tilde \beta }} = {\bf{K}}^{-1}{{\bf{H}}^T}{\bf{Y}},
\end{equation}
where, ${\bf{K}} = {({{\bf{H}}^T}{\bf{H}})^{ - 1}}$.
\subsection{Cascade Regression of Extreme Learning Machine}
\label{sec:Cascade Regression of Extreme Learning Machine}
A facial shape can be formed by a vector
${\bm{s}} = {[{r_1},{c_1}, \ldots ,{r_L},{c_L}]^\mathrm{T}}$ consisting of $L$ facial landmarks, where the $\left[ {{r_l},{c_l}} \right], l = 1,2, \ldots ,L$ are the 2D coordinates of the $l$-$th$ landmark. 
Cascade regression frameworks usually begin with an initial shape ${{\bm{s}}^{\bf{0}}}$, and progressively refine the shape to the ground-truth ${{\bm{s}}^{{*}}}$ via adding a shape increment $\Delta {\bm{s}}$ stage by stage. The $\Delta {\bm{s}}$ is estimated by regressing the shape-indexed feature around current shape estimate. The shape-indexed feature can be represented as ${\rm{f}}\left( {\bf{I}},\bm{s}  \right) \in \Re ^{f \times 1}$, where, $\bf{I}$ is the input image, $f$ is the dimensionality of the feature. The function ${\rm{f}}$ can be a learning-based mapping \cite{ren2014face,jourabloo2017pose} or a hand-crafted feature(e.g. SIFT \cite{xiong2013supervised}, Hog \cite{liu2019face}). Linear regression has been most favoured in various works based on cascade regression because its superior efficiency. However, it is not suitable for incremental learning framework because the shape variations are more and more complicated with increasing incremental training data. 
Therefore, we propose a cascade regression of extreme learning machine(CRELM). We following introduce the training procedure of CRELM.

Given a set of $N$  facial images ${\cal I} = \left\{ {{{\bf{I}}_i}} \right\}_{i = 1 \ldots N}$ and their corresponding ground-truth shapes ${\cal S} = \left\{ {{\bm{s}^{*}_{i}}} \right\}_{i = 1 \ldots N}$.
The set of shape increments can be calculated by $\delta \bm{s}_i^t =  \bm{s}_i^{t - 1} - \bm{s}_i^*$, where, $\bm{s}_i^{t - 1}$ is a $2l\times 1$ shape vector from $t-1$ stage.
For achieving a robust representation against illumination, we use SIFT features extracted from patches around the current shape of each stage. 
To decrease the training error for stage $t$, we can learn the stage-regressor ${\cal G}^{t}$ via minimizing the least-squares error function: 
\begin{equation}
\label{eq:01}
      \mathop {\arg \min }\limits_{{{\mathop{\cal {G}}\nolimits} ^t}} \sum\limits_{i = 1}^N {\left\| {\delta \bm{s}_i^t - {\mathop{\cal G}^t\nolimits} ({\mathop{\rm f}\nolimits} ({\bf{I}}, \bm{s}_i^{t - 1}))} \right\|_2^2}, 
\end{equation}
Let $\bm{x} = {\rm{f}}\left( \mathbf{I},\bm{s}  \right) \in \Re ^{f \times 1}$, ${\cal G}(\bm{x}) = \sum\limits_{k = 1}^K {{\bm{\beta }_k}} G({\bm{a}_k},{b_k},\bm{x})$ and $\Delta {\bf{S}} = {[\delta {\bm{s}_1}, \ldots ,\delta {\bm{s}_N}]^T}$, we can rewrite the Equation(\ref{eq:01}) as the format of ELM:
\begin{equation}
    \label{eq:CRELM_loss_function}
    \tilde {\bf{\beta }} ^{t}= \mathop {\arg \min }\limits_{{\bf{\beta }}^{t}} ||\Delta {\bf{S}}^{t} - {\bf{H}}^{t}{\bf{\beta}}^{t} ||_2^2,
\end{equation}
where. ${\bf{H}}^{t} = {\bf{H}}({\bm{a}^{t}_1}, \ldots ,{\bm{a}^{t}_K},{b^{t}_1}, \ldots ,{b^{t}_K},{\bm{x}_1}, \ldots ,{\bm{x}_N})$ is computed by Equation(\ref{eq:representation of H}) and we choose sigmoid function as the activation mapping $G(\bm{a},b,\bm{x}) = \frac{1}{{1 + \exp ( - (\bm{a} \cdot \bm{x + }b))}}$. 
The Equation (\ref{eq:CRELM_loss_function}) can be resolved by (\ref{eq:ELM solution_ordinary}), (\ref{eq:ELM solution nonsingular}). We represent the learned regressor for stage $t$ as ${{\cal G}^t} = [{\bm{a}^t}\bm{,}{b^t},{\tilde {\bf{\beta}} ^t},{\bf{K}}^{t}]$, where ${\bf{\tilde \beta }} = {\bf{K}}^{-1}{{\bf{H}}^T}\Delta {\bf{S}}$(for clearing, here, we omit $t$).
After learning the regressor, the training shapes for $t$ stage can be updated by:
\begin{equation}
\label{eq:02}
    \bm{s}_i^{t} = \bm{s}_i^{t - 1} - {{\cal G}^t}\left({\mathop{\rm f}\nolimits} \left( {{{\bf{I}}_i},\bm{s}_i^{t-1}} \right)\right).
\end{equation}
The training procedure is sequentially iterated until the average of the shape differences $\delta \bm{s}_i^t$ no longer decrease.
% Empirically speaking, the process usually invovles 4 or 5 cascade stages.

% For resolving the objective function Eq.\ref{eq:01}, linear regression is the most popular discriminant analysis in recent studies of face alignment \cite{ren2014face,xiong2013supervised,sanchez2016cascaded}. While it has a fast speed, the fitting capability is too weak to handle complex and non-linear relationship between shape variations and unconstrained environments. Thus, in our work, an efficient non-linear mapping is attempted to improve the accuracy of the whole model. In the next section, we will illustrate the proposed cascade regression of extreme learning machine.
% However, this way will inevitably increase the time-consuming of training process. Therefore, we adopt a parallel cascade framework, which can significantly save time cost and perform as accurately as the sequential cascade. In the section \ref{sec:Parallel Cascade of extreme learning machine}, we will illustrate the parallel cascade of extreme learning machine.

\begin{figure*}
\begin{center}
\begin{tabular}{c}
\includegraphics[height=6.2cm]{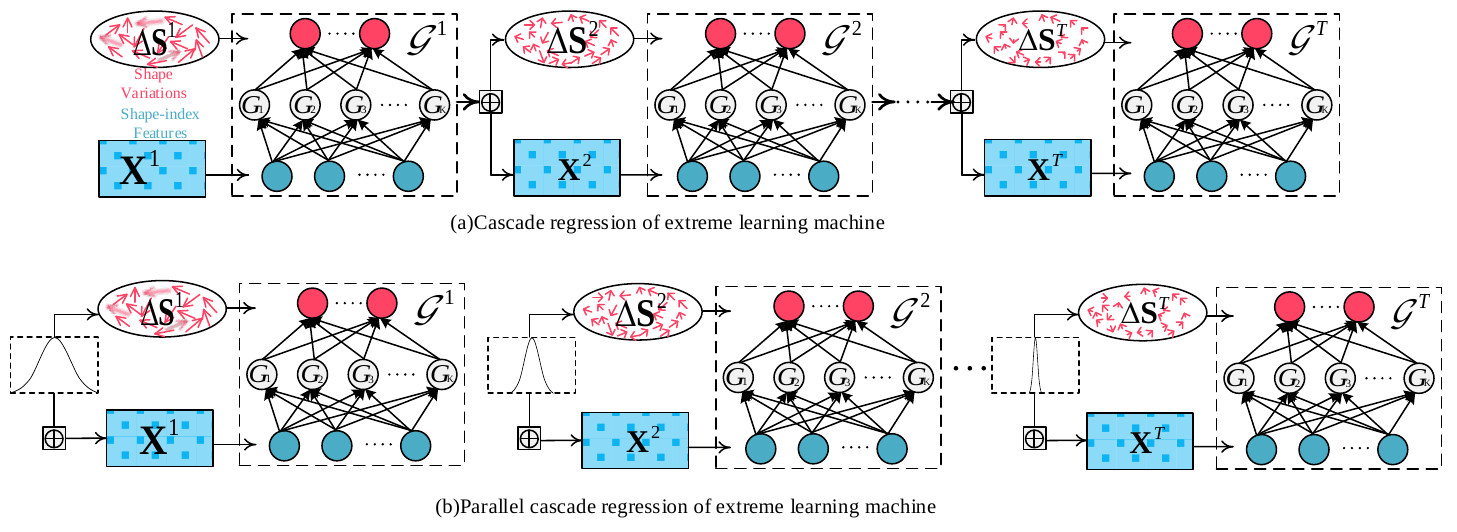}
\end{tabular}
\end{center}
\caption
{ \label{fig:03}
\textbf{The training procedures of CRELM and Par-CRELM}}
\end{figure*}

\subsection{Parallel Cascade Regression of Extreme learning machine}
\label{sec:Parallel Cascade Regression of Extreme learning machine}
In order to better approximate the non-linear relationship between the image features and shape variations, in Section~\ref{sec:Cascade Regression of Extreme Learning Machine}, we introduce an efficient non-linear mapping in cascade regression framework. However, this way will inevitably increase the time-consuming of training process.
Besides, it is observed that the sequential procedure involved in training CRELM is not suitable for the task of incremental learning. In CRELM, as shown in Figure~\ref{fig:03}(a), since the shape variations $\delta \bm{s}^{t}$ is totally depend on the outputting of previous stage, if new training data inputs, the entire cascade of regressors have to be retrained from the beginning. 
For example, if a set of new samples ${\cal S}_{new} = \left\{ {{\bm{s}^*}} \right\}_{i = 1 \ldots N'}$ have to be added, the first regressor ${\cal G}^{1}$ can be easily updated to ${\cal G}^{1}_{new}$ by utilizing the new shape variations $\delta \bm{s}^{1}_{new}=\bm{s}^{0}-\bm{s}^{*}_{new}$. However, the first regressor ${\cal G}^{1}_{new}$ has changed on once, the subsequent set of inputs $\bm{s}^{2}_{new} = \{ {\bm{s}^2},\bm{s}_{new}^2\}$ must be re-computed by progagating the entire augmented set $\bm{s}_{new}^1 = \{ {\bm{s}^1},\bm{s}_{new}^1\} $ through ${\cal G}^{1}_{new}$. In this case, all the regressors will retrained in sequence and
all previously trained samples must be reloaded, which is time consuming and extremely expensive. 

The authors of \cite{asthana2014incremental} pointed out that the shape variations $\delta \bm{s}_i^t$ at each stage can be approximated by a set of random perturbations drawn from a Gaussian distribution ${\cal N}({\mu ^{(t)}},{\Sigma ^{(t)}})$. the mean ${\mu ^{(t)}}$ and covariance ${\Sigma ^{(t)}}$ can be calculated by training the sequential cascade of regression. In addition, it has been verified in work \cite{asthana2014incremental} that
the training procedure based on the sampling strategy can deliver a similar testing 
accuracy as the sequential manner. 
Inspired by \cite{asthana2014incremental}, we adopt Monte Carlo sampling methodology to train a parallel cascade regression of extreme learning machine(Par-CRELM).
In Par-CRELM, the shape variations $\delta \bm{s}$ required for learning the cascade of regressors do not rely on previous stages and the training can be performed in parallel.  
In particular, firstly, we compute the the statics ${\sigma ^{(t)}} = ({\mu ^{(t)}},{\Sigma ^{(t)}})$ for shape variations $\left\{ {\Delta {\bm{s}_i}} \right\}_{i = 1}^N$ at each stage while training the
cascade of regressors using the proposed CRELM on the offline train set. Then, we train the Par-CRELM procedure, the shape variations for training the cascade of ELM regressors are drawn from the corresponding stage-distribution rather than the calcaulated from previous stage. 
We have shown the parallel process in Figure~\ref{fig:03}(b).
One advantage of this modification of CRELM is that the regressors of all stages can be learned in parallel with similar alignment accuracy as sequential training. Another advantage is that it provides a premise for fast incremental learning that will be showed in the Section \ref{sec:Incremental Cascade Regression of Extreme Leaning machine}. 

\subsection{Incremental Cascade Regression of Extreme Leaning machine}
\label{sec:Incremental Cascade Regression of Extreme Leaning machine}

After training Par-CRELM, the offline regressors $\left\{ {{{\cal G}^t}} \right\}_{t = 1}^T$ and the distribution ${\cal D}\left\{ {{\sigma ^{(1)}}, \ldots ,{\sigma ^{(T)}}} \right\}$ of shape variations are preserved. Here, we present the proposed incremental cascade regression of extreme learning machine(ICRELM) in detail.

Given a set of new training data ${{\cal I}_{new}} = {\left\{ {{{\bf{I}}_i}} \right\}_{i = 1 \ldots N'}}$ and ${\cal S}_{new} = \left\{ {{\bm{s}^*}} \right\}_{i = 1 \ldots N'}$, where $N'$ is the number of new samples. 
We record trained regressors and distributions as $\left\{ {{{\cal G}^t_{0}}} \right\}_{t = 1}^T$ and $\left\{{\cal D}^{t}_{0}\right\}_{t=1}^{T}$. For arbitrary stage, ${\cal G}^{t}_{0}$ contains learned parameters $[{\bm{a}^t}\bm{,}{b^t},{\tilde {\bf{\beta}}_{0} ^t},{\bf{K}}_{0}^{t}]$.
ICRELM aims to update the cascade of regressors $\left\{ {{{\cal G}^t_{0}}} \right\}_{t = 1}^T$( in which ${\bm{a}^t},{b^t}$ do not need to update)
in parallel using the new training data.
For stage $t$, let us randomly sample $N'$ shape variations $\{ \delta \bm{\tilde s}_i^t\} _{i = 1}^{N'}$ drawn from ${\cal N}({\mu ^{(t)}_{0}},{\Sigma ^{(t)}_{0}})$ for the new training images and extract the shape-index features $\{ \bm{x}_i^t\} _{i = 1}^{N'}$. The least-squares error function for all training data becomes: 
\begin{equation}
    \tilde{\bf{\beta }}_{new}^t = \mathop {\arg \min }\limits_{{{\bf{\beta }}^t_{new}}} \left\| {\left[ \begin{array}{l}
\Delta {{\bf{S}}^t_{0}}\\
\Delta {\bf{S}}_{new}^t
\end{array} \right] - \left[ \begin{array}{l}
{{\bf{H}}^t_{0}}\\
{\bf{H}}_{new}^t
\end{array} \right]{{\bf{\beta }}^t_{new}}} \right\|_2^2,
\end{equation}
where, $\Delta {\bf{S}}^{t}_{new} = {[\delta {\bm{\tilde s}_1}, \ldots ,\delta {\bm{\tilde s}_{N'}}]^T}$, ${\bf{H}}^{t}_{new} = {\bf{H}}({\bm{a}^{t}_1}, \ldots ,{\bm{a}^{t}_K},{b^{t}_1}, \ldots ,{b^{t}_K},{\bm{x}_1}, \ldots ,\bm{x}_{N'})$ is computed by Equation(\ref{eq:representation of H}) using the trained parameters $[{\bm{a}^t}\bm{,}{b^t}]$ of regressor ${\cal G}^{t}_{0}$. Then, the output weight $\bf{\beta}$ can be calculated by Equation(\ref{eq:ELM solution nonsingular}):
\begin{equation}
    \label{eq: update beta}
    {\bf{\tilde \beta }}_{new}^t = {{\bf{K}}_{new}^{t}}^{ - 1}{\left[ \begin{array}{l}
{{\bf{H}}^t_{0}}\\
{\bf{H}}_{new}^t
\end{array} \right]^T}\left[ \begin{array}{l}
\Delta {{\bf{S}}^t_{0}}\\
\Delta {\bf{S}}_{new}^t
\end{array} \right].
\end{equation}
where
\begin{equation}
\label{eq: update K}
    {{\bf{K}}_{new}^{t}} = {{\bf{K}}_0^{t}} + {{\bf{H}}^{t}_{new}}^T{{\bf{H}}^{t}_{new}}.
\end{equation}
Referring to \cite{liang2006fast}, we can update the ${\bf{\tilde \beta }}_{new}^t$ via:
\begin{equation}
\label{eq: compute betanew}
    {\bf{\tilde \beta }}_{new}^t = {\bf{\tilde \beta }}_0^t + {\bf{K}}_{new}^{ - 1}{{\bf{H}}{_{new}^t}^T}(\Delta {\bf{S}}_{new}^t - {\bf{H}}_{new}^t{\bf{\tilde \beta }}_0^t)
\end{equation}

This way, a cascade of regressors can be updated in parallel. The complete training procedure of ICRELM is described in Algorithm~\ref{algo:ICRELM Update Procedure}.
\begin{algorithm}
\caption{ICRELM Update Procedure}
\label{algo:ICRELM Update Procedure}
    \begin{algorithmic}[1]
        \REQUIRE
        ${{\cal I}_{new}}$, ${{\cal S}_{new}}$, $\left\{ {{{\cal G}^t_{0}}}=[{\bm{a}^t}\bm{,}{b^t},{\tilde {\bf{\beta}}_{0} ^t},{\bf{K}}_{0}^{t}] \right\}_{t = 1}^T$, $\left\{{\cal D}^{t}_{0}\right\}_{t=1}^{T}$, $T$ stages, $N'$ number of new samples.
        \ENSURE 
        updated regressors ${\cal G}_{new}^t = [{\bm{a}^t},{b^t},{\bf{\tilde \beta }}_{new}^t,K_{new}^t]$, $t = 1, \ldots ,T$.
        \STATE {\bf{Parallel for}} $t = 1 \to T$
        \STATE 
        \begin{itemize}
            \item[] Get $N'$ samples $\{ \delta \bm{s}_i^t\} _{i = 1}^{N'}$ from distribution ${\sigma ^{(t)}}$
        \end{itemize}
        \STATE 
        \begin{itemize}
            \item[] Extract index-shape features $\{ \bm{x}_i^t = {\mathop{\rm f}\nolimits} ({\bf{I}},\bm{s}_i^t)\} _{i = 1}^{N'}$ 
        \end{itemize}
        
        \STATE
        \begin{itemize}
            \item[] Generate $\Delta {\bf{S}} \in {\Re ^{N' \times f}}$ and ${\bf{X}} \in {\Re ^{N' \times 2l}}$
        \end{itemize}
        
        \STATE 
        \begin{itemize}
            \item[] Compute ${\bf{H}}_{new}^t$ using Equation(\ref{eq:representation of H})
        \end{itemize}

        \STATE 
        \begin{itemize}
            \item[] Update ${\bf{K}}_{new}^t$ using Equation(\ref{eq: update K}) and ${\bf{\tilde \beta }}_{new}^t$ using Equation(\ref{eq: compute betanew})
        \end{itemize}
        \STATE {\bf{End for}}

    \end{algorithmic}
\end{algorithm}

\section{Experiments}
The experiments for face alignment will be presented in two parts . The first part is to evaluate the accuracy of the model which constantly updated by ICRELM with continuous batches of new training data. The second part investigates the static models trained by proposed Par-CRELM and other state-of-the-art methods on public datasets. First of all, We briefly introduce the three datasets used in the experiments of face alignment and evaluation criteria for them.

\subsection{Implementation Details}
\subsubsection{Datasets}
\begin{itemize}
    \item  \emph{LFPW} (29 landmarks) \cite{le2012interactive} is collected from the web including 1000 training and 300 test images. However some URLs are invalid, we only use 798 training and 221 test images. The images exhibit large variations in pose, occlusion, facial expression, and illumination.
    
    \item {\emph{HELEN}} (68 landmarks) \cite{belhumeur2013localizing} contains 2,330 high-resolution web images which are divided into training and test sets. Two sets have 2000 and 300 images respectively.
    
    \item {\emph{300-W}} (68 landmarks) \cite{sagonas2013300} is collected from existing datasets including LFPW, HELLEN, AFW and a challenging dataset called IBUG. We follow the same division in \cite{ren2014face, shi2018face}, specifically, the training set is made up by the training samples of HELEN, the training samples of LFPW, and AFW, with 3148 images in total. According to the difficulty of alignment, the test set is grouped into Common(the test samples from LFPW and HELEN, total 554 images) and Challenging(IBUG, total 135 images) sets.
\end{itemize}
 
Since these datasets provide prescribed face bounding boxes, we do not use any face detectors and thus no face are missed during testing.

\subsubsection{Standard Evaluation Protocols}
We adopt two types of comparisons with the value of average error and curve of cumulative error for evaluation. They are prescribed as following:
\begin{itemize}
    \item {\emph{Average error:}} following almost works, we leverage
    the standard landmarks mean error normalized by inter-pupil distance. It can be computed by $\frac{{\sum\limits_{l = 1}^L {||{p_l} - } p_l^*|{|_2}}}{{L \cdot ||p_{leye}^* - p_{reye}^*|{|_2}}}$, where, ${{p_l}}$ and ${p_l^*}$ denote the $l$-$th$ landmark coordinates of estimated and ground-truth facial landmark positions respectively, the ${p_{leye}^*}$ and ${p_{reye}^*}$ denote the inter-pupil distance. We report the error averaged over all annotated landmarks from each testing database. For clarity, we omit the notation $\%$ in the report result.
    \item{\emph{CED curve:}} we also draw the cumulative error distribution curve of errors can be computed by the equation: $CED = \frac{{{N_k} \le e}}{N}$, where the numerator denotes the number of samples on which the error less than the error $e$.
    
\end{itemize}

\subsubsection{Settings}
In the feature learning, we extracted a SIFT \cite{lowe1999object} descriptor on $32 \times 32$ local patches for each landmark. We set the number of hidden nodes as 500, 1000, and 1800 for LFPW, HELEN, 300-W datasets respectively. Following almost face alignment models, we fixed the number of cascade stages to $T=4$.

\subsection{Validity of Online Learning}
\begin{figure}[hbt]

\centering
\subfigure[LFPW Test set]{
    \includegraphics[width=3in]{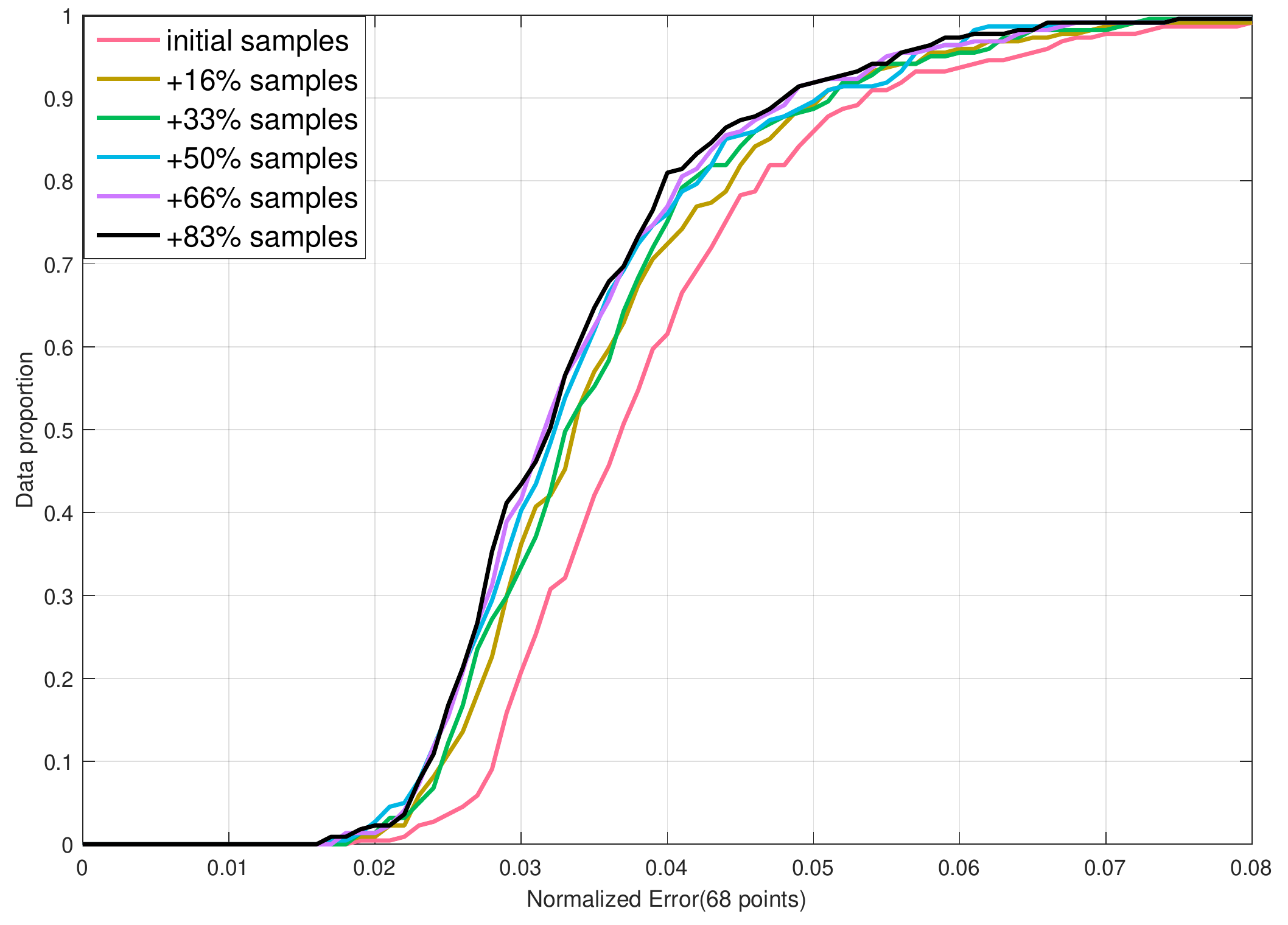}
}
\subfigure[HELEN Test set]{
\includegraphics[width=3in]{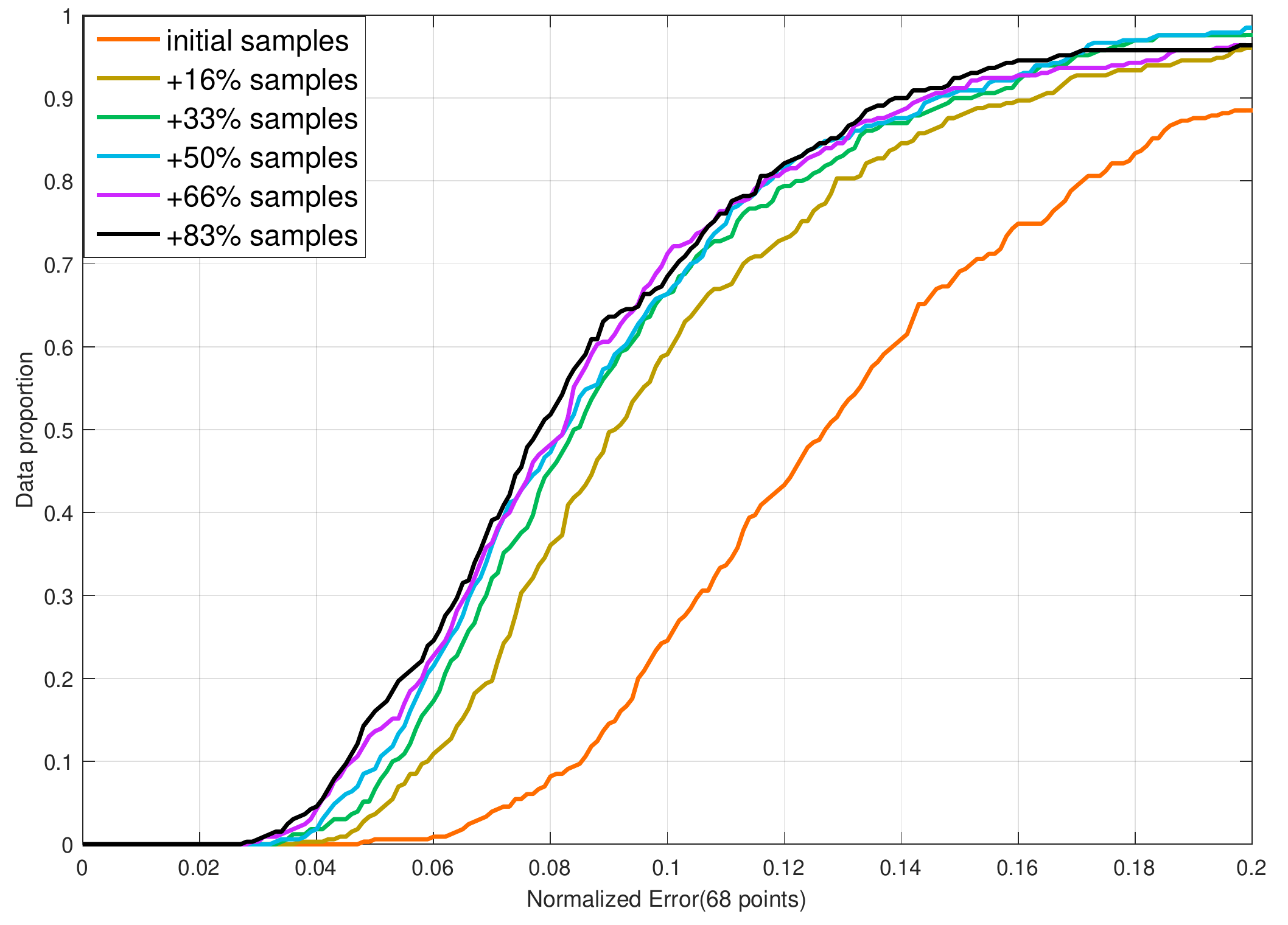}}
\subfigure[300-W Test set]{
\includegraphics[width=3in]{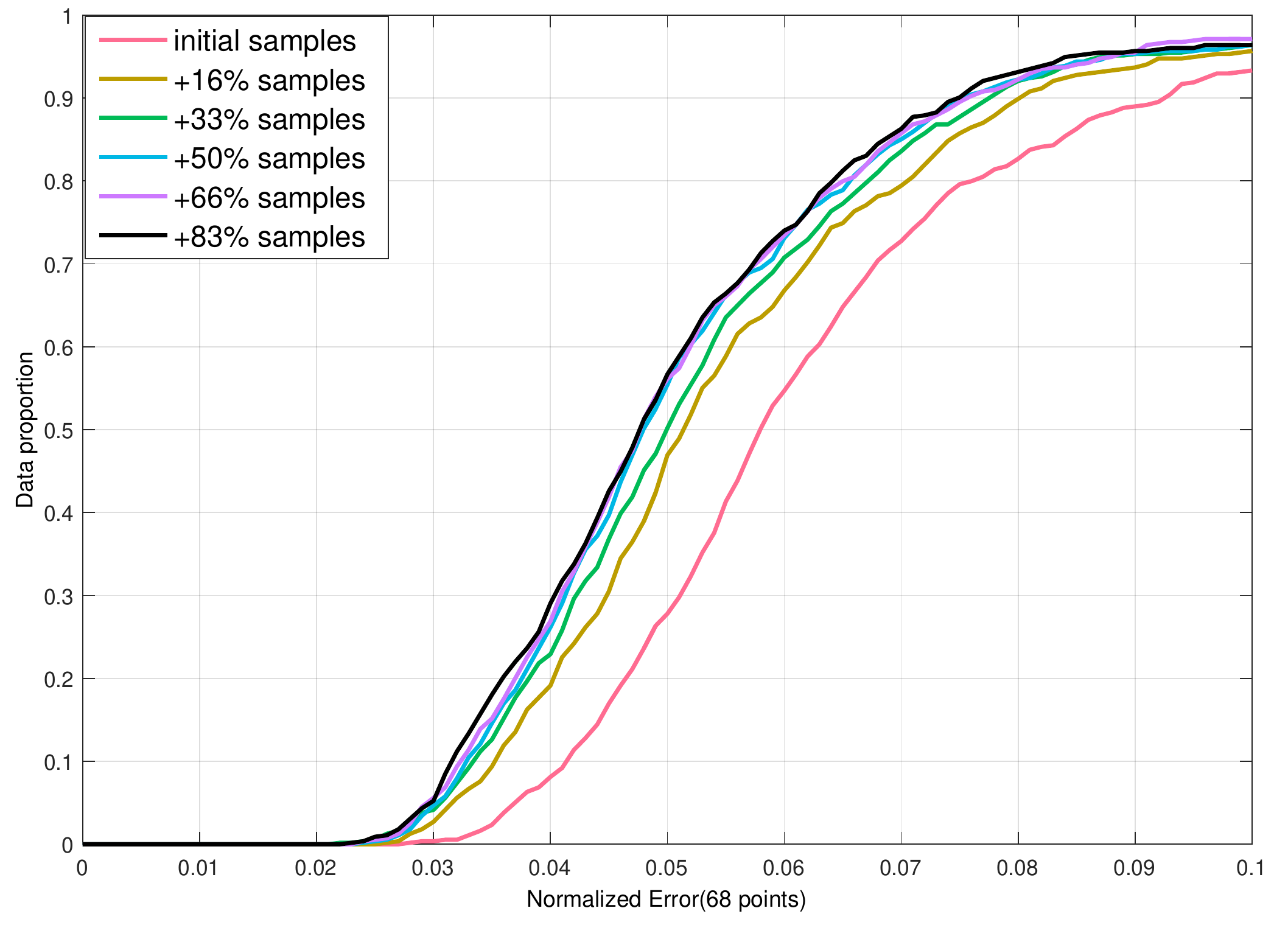}
}
\caption{ICRELM results on different datasets}
\label{fig:validity of ICRELM}
\end{figure}

This experiment aims to validate the utility of the ICRELM(Section~\ref{sec:Incremental Cascade Regression of Extreme Leaning machine}) when the new data batches continuously arrive. For this purpose, we have designed the following experiments for LFPW, HELEN, and 300-W datasets. Each dataset was partitioned equally into 6 batches. We used Par-CRELM(Section~\ref{sec:Parallel Cascade Regression of Extreme learning machine}) to train a offline model on the first batch as the baseline.
Then, we employed ICRELM to continuously update the generic model batch-by-batch. 
From Figure~\ref{fig:validity of ICRELM}, we can observe that a consistent increase in the face alignment accuracy as the online model is incrementally updated with new batch of training data. The worst curve is produced by offline model. It is inevitable that the model trained with small samples tend to have poor generalization and robustness. The other curves are generated after adding $16$\%, $33$\%, $50$\%, $66$\%, and $83$\% samples in batches from corresponding dataset. The curves illustrate that ICRELM method can update the model effectively when necessary and achieve higher and higher accuracy as training data increases. It is also observed in these curves that the accuracy of the updated model is no longer significantly raised after the data increases by 66\%. It is because the difference between training sets is getting smaller and smaller, and the generalization capability of the model tends to be stable. 
It is notable that the ICRELM has a very fast speed to update the generic model. ICRELM is implemented in Matlab and ran on an Intel Single Core i5-4570@3.2GHz CPU at over 110fps, 33, and 24fps on LFPW, HELEN, and 300-W dataset respectively.

\subsection{Comparison with static models}
\label{Comparison with static models} 
Offline model is the foundation of online learning. In order to validate the capability of the model trained by Par-CRELM on different datasets, we compare its results with existing state-of-the-art methods including CNN-based frameworks \cite{shi2018face,shi2014deep,r_35,r_65,jourabloo2017pose,liu2017dense,zhang2018exemplar}, 3D-based model\cite{r_67} and various cascade regressions(CRs). These results have been reported in the Table~\ref{tab:summary_listing}. On the LFPW dataset(Table~\ref{tab:firsttable}), as we can seen, Par-CRELM outperforms all the listed CR-based methods.
Meanwhile, Par-CRELM can also generate a competitive result compared with CNN-based architectures. For each dataset, the Par-CRELM is lower than DR. Except for structural difference, one possible reason is that DR uses more cascaded iterations than Par-CRELM. Correspondingly, more time was consumed than ours both in training or testing. On the HELEN dataset(Table~\ref{tab:secondtable}), our approach has a more accurate result than CFAN, MTCNN. Specifically, CFAN maps the local features to the shape space by utilizing deep auto-encoder networks, MTCNN applys the Multi-task convolution network in face alignment. 
The result shows that the Par-CRELM has a superior learning capability on smaller scaled data. Conversely, the CNN-based methods are prone to be restricted by the size of the dataset. Whereas our approach offers an advantage on HELLEN dataset. On the 300-W Common subset(Table~\ref{tab:thirdtable}), the result of Par-CRELM is superior to almost cascade regression methods and certain CNN frameworks like DR-Seq, SCNN DeFA and GECSAN, but lower than the methods including DR, Deep Regression. This dataset has a larger data size than LFPW and HELEN do, which provides sufficient discriminative information for CNN-based methods learning features. This may be more conductive for the CNN-based methods producing better results. Unfortunately, Par-CRELM performs unsatisfactorily in Challenging subset. In contrast, the CNN-based methods, such as SCNN, DeFA, and CECSAN, have highly accuracies. The reason is that they can learn more adequate feature via tuning by supervised information, which is essential for good performance on very challenging samples. 
While the static model trained by Par-ELM has a poor predictive capability to handle these challenging situations, it can be tuned with challenging samples to improve its localizing ability.

\begin{table*}
\newcommand{\tabincell}[2]{\begin{tabular}{@{}#1@{}}#2\end{tabular}}
\footnotesize
\caption{\textbf{Results on the averaged errors with the state-of-the-art approaches (the top 2 results for each dataset are marked in bold). The results of all methods have reported in the original papers or related literatures.}} 
\label{tab:summary_listing}
% \centering {RESULTS ON THE AVERAGED ERRORS WITH THE STATE-OF-THE-ART APPROACHES(THE TOP 2 RESULTS FOR EACH DATA SET ARE MARKED IN BOLD). THE RESULTS OF ALL METHODS HAVE REPORTED IN THE ORIGINAL PAPERS OR RELATED LITERATURES}\\
~\\
\centering  
\subtable[\textbf{LFPW}]{  
\begin{tabular}{ccc} %% this creates two columns
%% |l|l| to left justify each column entry
%% |c|c| to center each column entry
%% use of \rule[]{}{} below opens up each row
\hline\hline
\rule[-1ex]{0pt}{3.5ex} \tabincell{c}{Methods\\~} & \tabincell{c}{Error\\~} 
% & \tabincell{c}{FPS\\~} 
\\
\hline
\rule[-1ex]{0pt}{3.5ex} OC \cite{r_63} & 5.07 \\
\rule[-1ex]{0pt}{3.5ex} CE \cite{belhumeur2013localizing} & 3.99 \\
\rule[-1ex]{0pt}{3.5ex} EGM \cite{r_60} & 3.98 \\
\rule[-1ex]{0pt}{3.5ex} RFLD \cite{r_62} & 3.93 \\
\rule[-1ex]{0pt}{3.5ex} -              & -    \\
\rule[-1ex]{0pt}{3.5ex} PCPR \cite{r_31} & 3.50 \\
\rule[-1ex]{0pt}{3.5ex} SDM \cite{xiong2013supervised} & 3.49 \\
\rule[-1ex]{0pt}{3.5ex}  DR-Seq \cite{shi2018face} & 3.90 \\
\rule[-1ex]{0pt}{3.5ex} DR-SDM \cite{shi2018face} & 3.40 \\
\rule[-1ex]{0pt}{3.5ex} ESR \cite{cao2014face} & 3.47 \\
\rule[-1ex]{0pt}{3.5ex} LBF \cite{ren2014face} & 3.35 \\
\rule[-1ex]{0pt}{3.5ex} -    & -    \\
\rule[-1ex]{0pt}{3.5ex} Deep Reg
\cite{shi2014deep} & 3.45  \\
\rule[-1ex]{0pt}{3.5ex} RSR \cite{r_61} & 3.34 \\
\rule[-1ex]{0pt}{3.5ex} cGPRT\cite{lee2015face} & 3.51  \\
\rule[-1ex]{0pt}{3.5ex} -    & - \\
\rule[-1ex]{0pt}{3.5ex} DR \cite{shi2018face} & \textbf{3.31}  \\
\rule[-1ex]{0pt}{3.5ex} -    & -    \\
\rule[-1ex]{0pt}{3.5ex} -    & -    \\
\rule[-1ex]{0pt}{3.5ex} -    & -    \\
\rule[-1ex]{0pt}{3.5ex} -    & -    \\
\rule[-1ex]{0pt}{3.5ex} -    & -    \\

\hline\hline
\rule[-1ex]{0pt}{3.5ex} Ours & \textbf{3.32}  \\
\hline\hline
\end{tabular}
\label{tab:firsttable}  
}
%\qquad  
\subtable[\textbf{HELEN}]{          
      \begin{tabular}{cc} %% this creates two columns
%% |l|l| to left justify each column entry
%% |c|c| to center each column entry
%% use of \rule[]{}{} below opens up each row
\hline\hline
\rule[-1ex]{0pt}{3.5ex} \tabincell{c}{Method\\~} & \tabincell{c}{Error\\~} \\
\hline
\rule[-1ex]{0pt}{3.5ex}  -             & -   \\
\rule[-1ex]{0pt}{3.5ex}  -             & -   \\
\rule[-1ex]{0pt}{3.5ex}  -             & -   \\
\rule[-1ex]{0pt}{3.5ex}  -             & -   \\
\rule[-1ex]{0pt}{3.5ex} Zhu et.al \cite{r_47} & 8.16 \\
\rule[-1ex]{0pt}{3.5ex} PCPR \cite{r_31} & 5.93 \\
\rule[-1ex]{0pt}{3.5ex} SDM \cite{xiong2013supervised} & 5.50 \\
\rule[-1ex]{0pt}{3.5ex} DR-Seq\cite{shi2018face} & 6.47 \\
\rule[-1ex]{0pt}{3.5ex} DR-SDM\cite{shi2018face} & 5.40 \\
\rule[-1ex]{0pt}{3.5ex}  -             & -   \\
\rule[-1ex]{0pt}{3.5ex}  -             & -   \\
\rule[-1ex]{0pt}{3.5ex} CFAN \cite{r_35} & 5.53 \\
\rule[-1ex]{0pt}{3.5ex}  -             & -   \\
\rule[-1ex]{0pt}{3.5ex}  -             & -   \\
\rule[-1ex]{0pt}{3.5ex}  -             & -   \\
\rule[-1ex]{0pt}{3.5ex} MTCNN \cite{r_65} & 5.49 \\
\rule[-1ex]{0pt}{3.5ex} DR \cite{shi2018face} & \textbf{5.09} \\
\rule[-1ex]{0pt}{3.5ex}  -             & -   \\
\rule[-1ex]{0pt}{3.5ex}  -             & -   \\
\rule[-1ex]{0pt}{3.5ex}  -             & -   \\
\rule[-1ex]{0pt}{3.5ex}  -             & -   \\
\rule[-1ex]{0pt}{3.5ex}  -             & -   \\

\hline\hline
\rule[-1ex]{0pt}{3.5ex} Ours & \textbf{5.37} \\
\hline\hline
\end{tabular} 
\label{tab:secondtable}  
}  
%\qquad  
\subtable[\textbf{300-W}]{          
\begin{tabular}{ccc} %% this creates two columns
%% |l|l| to left justify each column entry
%% |c|c| to center each column entry
%% use of \rule[]{}{} below opens up each row
\hline\hline
\rule[-1ex]{0pt}{3.5ex}  \tabincell{c}{Method\\~}	& \tabincell{c}{Common\\Subset} & \tabincell{c}{Challenging\\Subset}\\
\hline
\rule[-1ex]{0pt}{3.5ex}  -     &-      &-  \\
\rule[-1ex]{0pt}{3.5ex}  -     &-      &-  \\
\rule[-1ex]{0pt}{3.5ex}  -     &-      &-  \\
\rule[-1ex]{0pt}{3.5ex}  -     &-      &-  \\
\rule[-1ex]{0pt}{3.5ex}  Zhu et.al \cite{r_47}	&8.22	&18.33 \\
\rule[-1ex]{0pt}{3.5ex}  PCPR \cite{r_31}	&6.18	&17.26 \\
\rule[-1ex]{0pt}{3.5ex}  SDM \cite{xiong2013supervised}	&5.70	&15.40  \\
\rule[-1ex]{0pt}{3.5ex}  DR-Seq\cite{shi2018face}
&5.44   &17.6\\
\rule[-1ex]{0pt}{3.5ex}  DR-SDM\cite{shi2018face}
&4.67   &14.30\\
\rule[-1ex]{0pt}{3.5ex}  ESR \cite{cao2014face}	&5.28	&17.00 \\
\rule[-1ex]{0pt}{3.5ex}  LBF \cite{ren2014face}	&4.95	&11.98 \\
\rule[-1ex]{0pt}{3.5ex}  CFAN \cite{r_35}	&5.50	&- \\
\rule[-1ex]{0pt}{3.5ex} Deep Reg \cite{shi2014deep}
&4.51   &13.80\\
\rule[-1ex]{0pt}{3.5ex}  -     &-      &-  \\
\rule[-1ex]{0pt}{3.5ex}  -     &-      &-  \\
\rule[-1ex]{0pt}{3.5ex}  -     &-      &-  \\
\rule[-1ex]{0pt}{3.5ex}  DR \cite{shi2018face}
&\textbf{4.35}   &13.3\\
\rule[-1ex]{0pt}{3.5ex}  3DDFA \cite{r_67}
&6.15   &10.59\\
\rule[-1ex]{0pt}{3.5ex}  SCNN \cite{jourabloo2017pose}	&5.43	&9.88 \\
\rule[-1ex]{0pt}{3.5ex}  3DDFA+SDM\cite{r_67}
&5.53   &9.56\\
\rule[-1ex]{0pt}{3.5ex}  DeFA \cite{liu2017dense}
&5.37   &9.38\\
\rule [-1ex]{0pt}{3.5ex}
GECSAN \cite{zhang2018exemplar}  
&5.42   &11.80  \\
\hline\hline
\rule[-1ex]{0pt}{3.5ex}  Ours	& \textbf{5.10}	&13.87 \\
\hline\hline
\end{tabular}
\label{tab:thirdtable}  
} 
\end{table*}  

\subsection{Conclusion and future works}

In this paper, we have proposed a incremental learning framework for face alignment, coined incremental cascade regression(ICR), which includes offline and online training procedures. A cascade regression of extreme learning machine is first introduced and its parallel version is developed to train a offline model. Then, we present an efficient method to incrementally update a trained model to make it more generalizable or specific.
The experimental results demonstrate the validity of online learning.  
Using our MATLAB implementation, the entire incremental learning procedure takes over 110fps, 33fps, 24fps on LFPW, HELEN, and 300-W dataset respectiely, on an Intel Single Core i5-4570@3.2GHz CPU computer. Possible future works include replacing the hand-crafted SIFT features with deep features \cite{esmaeilpour2019robust,wu2019cycle} and new optimization strategy\cite{wang2016iterative,wang2018multiview}.

% \subsection{Acknowledgement}
% This study was funded by National Natural Science Foundation of People's Republic of China (61672130, 61602082, 91648205), the  National Key Scientific Instrument and Equipment Development Project (No. 61627808), the Development of Science and Technology of Guangdong Province Special Fund Project Grants (No. 2016B090910001).

\bibliographystyle{ieeetr}
\bibliography{refs}

\begin{thebibliography}{10}

\bibitem{wu2019cross}
L.~Wu, R.~Hong, Y.~Wang, and M.~Wang, ``Cross-entropy adversarial view
  adaptation for person re-identification,'' {\em IEEE Transactions on Circuits
  and Systems for Video Technology}, 2019.

\bibitem{qiu2018eabs}
T.~Qiu, R.~Qiao, and D.~O. Wu, ``Eabs: An event-aware backpressure scheduling
  scheme for emergency internet of things,'' {\em IEEE Transactions on Mobile
  Computing}, vol.~17, no.~1, pp.~72--84, 2018.

\bibitem{cootes2001active}
T.~F. Cootes, G.~J. Edwards, and C.~J. Taylor, ``Active appearance models,''
  {\em IEEE Transactions on Pattern Analysis \& Machine Intelligence}, no.~6,
  pp.~681--685, 2001.

\bibitem{tzimiropoulos2014gauss}
G.~Tzimiropoulos and M.~Pantic, ``Gauss-newton deformable part models for face
  alignment in-the-wild,'' in {\em Proceedings of the IEEE Conference on
  Computer Vision and Pattern Recognition}, pp.~1851--1858, 2014.

\bibitem{liu2009discriminative}
X.~Liu, ``Discriminative face alignment,'' {\em IEEE Transactions on Pattern
  Analysis and Machine Intelligence}, vol.~31, no.~11, pp.~1941--1954, 2009.

\bibitem{xiong2013supervised}
X.~Xiong and F.~De~la Torre, ``Supervised descent method and its applications
  to face alignment,'' in {\em Proceedings of the IEEE conference on computer
  vision and pattern recognition}, pp.~532--539, 2013.

\bibitem{shi2018face}
B.~Shi, X.~Bai, W.~Liu, and J.~Wang, ``Face alignment with deep regression,''
  {\em IEEE transactions on neural networks and learning systems}, vol.~29,
  no.~1, pp.~183--194, 2018.

\bibitem{lee2015face}
D.~Lee, H.~Park, and C.~D. Yoo, ``Face alignment using cascade gaussian process
  regression trees,'' in {\em Proceedings of the IEEE Conference on Computer
  Vision and Pattern Recognition}, pp.~4204--4212, 2015.

\bibitem{asthana2014incremental}
A.~Asthana, S.~Zafeiriou, S.~Cheng, and M.~Pantic, ``Incremental face alignment
  in the wild,'' in {\em Proceedings of the IEEE conference on computer vision
  and pattern recognition}, pp.~1859--1866, 2014.

\bibitem{tzimiropoulos2015project}
G.~Tzimiropoulos, ``Project-out cascaded regression with an application to face
  alignment,'' in {\em Proceedings of the IEEE Conference on Computer Vision
  and Pattern Recognition}, pp.~3659--3667, 2015.

\bibitem{cui2018recurrent}
Z.~Cui, S.~Xiao, Z.~Niu, S.~Yan, and W.~Zheng, ``Recurrent shape regression,''
  {\em IEEE transactions on pattern analysis and machine intelligence}, 2018.

\bibitem{ren2014face}
S.~Ren, X.~Cao, Y.~Wei, and J.~Sun, ``Face alignment at 3000 fps via regressing
  local binary features,'' in {\em Proceedings of the IEEE Conference on
  Computer Vision and Pattern Recognition}, pp.~1685--1692, 2014.

\bibitem{cao2014face}
X.~Cao, Y.~Wei, F.~Wen, and J.~Sun, ``Face alignment by explicit shape
  regression,'' {\em International Journal of Computer Vision}, vol.~107,
  no.~2, pp.~177--190, 2014.

\bibitem{dollar2010cascaded}
P.~Doll{\'a}r, P.~Welinder, and P.~Perona, ``Cascaded pose regression,'' in
  {\em 2010 IEEE Computer Society Conference on Computer Vision and Pattern
  Recognition}, pp.~1078--1085, IEEE, 2010.

\bibitem{jin2017face}
X.~Jin and X.~Tan, ``Face alignment in-the-wild: A survey,'' {\em Computer
  Vision and Image Understanding}, vol.~162, pp.~1--22, 2017.

\bibitem{wu2019cycle}
L.~Wu, Y.~Wang, and L.~Shao, ``Cycle-consistent deep generative hashing for
  cross-modal retrieval,'' {\em IEEE Transactions on Image Processing},
  vol.~28, no.~4, pp.~1602--1612, 2019.

\bibitem{jourabloo2017pose}
A.~Jourabloo, M.~Ye, X.~Liu, and L.~Ren, ``Pose-invariant face alignment with a
  single cnn,'' in {\em Proceedings of the IEEE International Conference on
  Computer Vision}, pp.~3200--3209, 2017.

\bibitem{wang2017effective}
Y.~Wang, X.~Lin, L.~Wu, and W.~Zhang, ``Effective multi-query expansions:
  Collaborative deep networks for robust landmark retrieval,'' {\em IEEE
  Transactions on Image Processing}, vol.~26, no.~3, pp.~1393--1404, 2017.

\bibitem{wu20193}
L.~Wu, Y.~Wang, L.~Shao, and M.~Wang, ``3-d personvlad: Learning deep global
  representations for video-based person reidentification,'' {\em IEEE
  transactions on neural networks and learning systems}, 2019.

\bibitem{sanchez2018functional}
E.~S{\'a}nchez-Lozano, G.~Tzimiropoulos, B.~Martinez, F.~De~la Torre, and
  M.~Valstar, ``A functional regression approach to facial landmark tracking,''
  {\em IEEE transactions on pattern analysis and machine intelligence},
  vol.~40, no.~9, pp.~2037--2050, 2018.

\bibitem{liang2006fast}
N.-Y. Liang, G.-B. Huang, P.~Saratchandran, and N.~Sundararajan, ``A fast and
  accurate online sequential learning algorithm for feedforward networks,''
  {\em IEEE Transactions on neural networks}, vol.~17, no.~6, pp.~1411--1423,
  2006.

\bibitem{huang2012extreme}
G.~B. Huang, H.~Zhou, X.~Ding, and R.~Zhang, ``Extreme learning machine for
  regression and multiclass classification,'' {\em IEEE Transactions on
  Systems, Man, and Cybernetics, Part B (Cybernetics)}, vol.~42, no.~2,
  pp.~513--529, 2012.

\bibitem{sung2009adaptive}
J.~Sung and D.~Kim, ``Adaptive active appearance model with incremental
  learning,'' {\em Pattern recognition letters}, vol.~30, no.~4, pp.~359--367,
  2009.

\bibitem{levey2000sequential}
A.~Levey and M.~Lindenbaum, ``Sequential karhunen-loeve basis extraction and
  its application to images,'' {\em IEEE Transactions on Image processing},
  vol.~9, no.~8, pp.~1371--1374, 2000.

\bibitem{zhang2018robust}
J.~Zhang, H.~Wang, and Y.~Ren, ``Robust tracking via weighted online extreme
  learning machine,'' {\em Multimedia Tools and Applications}, pp.~1--25, 2018.

\bibitem{liu2019face}
C.~Liu, L.~Feng, H.~Wang, and B.~Wu, ``Face alignment via multi-regressors
  collaborative optimization,'' {\em IEEE Access}, vol.~7, pp.~4101--4112,
  2019.

\bibitem{le2012interactive}
V.~Le, J.~Brandt, Z.~Lin, L.~Bourdev, and T.~S. Huang, ``Interactive facial
  feature localization,'' in {\em European conference on computer vision},
  pp.~679--692, Springer, 2012.

\bibitem{belhumeur2013localizing}
P.~N. Belhumeur, D.~W. Jacobs, D.~J. Kriegman, and N.~Kumar, ``Localizing parts
  of faces using a consensus of exemplars,'' {\em IEEE transactions on pattern
  analysis and machine intelligence}, vol.~35, no.~12, pp.~2930--2940, 2013.

\bibitem{sagonas2013300}
C.~Sagonas, G.~Tzimiropoulos, S.~Zafeiriou, and M.~Pantic, ``300 faces
  in-the-wild challenge: The first facial landmark localization challenge,'' in
  {\em Proceedings of the IEEE International Conference on Computer Vision
  Workshops}, pp.~397--403, 2013.

\bibitem{lowe1999object}
D.~G. Lowe {\em et~al.}, ``Object recognition from local scale-invariant
  features.,'' in {\em iccv}, vol.~99, pp.~1150--1157, 1999.

\bibitem{shi2014deep}
B.~Shi, X.~Bai, W.~Liu, and J.~Wang, ``Deep regression for face alignment,''
  {\em arXiv preprint arXiv:1409.5230}, 2014.

\bibitem{r_35}
J.~Zhang, S.~Shan, M.~Kan, and X.~Chen, ``Coarse-to-fine auto-encoder networks
  (cfan) for real-time face alignment,'' in {\em European Conference on
  Computer Vision}, vol.~8690, pp.~1--16, 2014.

\bibitem{r_65}
Y.~Sun, X.~Zhang, and C.~Li, ``Multi-task convolution network for face
  alignment,'' in {\em Journal of Physics: Conference Series}, vol.~887,
  p.~012079, 2017.

\bibitem{liu2017dense}
Y.~Liu, A.~Jourabloo, W.~Ren, and X.~Liu, ``Dense face alignment,'' in {\em
  Proc. IEEE Int. Conf. Comput. Vis. Workshops}, pp.~1619--1628, 2017.

\bibitem{zhang2018exemplar}
J.~Zhang and H.~Hu, ``Exemplar-based cascaded stacked auto-encoder networks for
  robust face alignment,'' {\em Computer Vision and Image Understanding},
  vol.~171, pp.~95--103, 2018.

\bibitem{r_67}
X.~Zhu, Z.~Lei, and S.~Z. Li, ``Face alignment in full pose range: A 3d total
  solution,'' {\em IEEE Transactions on Pattern Analysis and Machine
  Intelligence}, 2017.

\bibitem{r_63}
G.~Ghiasi and C.~C. Fowlkes, ``Occlusion coherence: Localizing occluded faces
  with a hierarchical deformable part model,'' in {\em Proceedings of the IEEE
  Conference on Computer Vision and Pattern Recognition}, pp.~2385--2392, 2014.

\bibitem{r_60}
F.~Zhou, J.~Brandt, and Z.~Lin, ``Exemplar-based graph matching for robust
  facial landmark localization,'' in {\em Computer Vision (ICCV), 2013 IEEE
  International Conference on}, pp.~1025--1032, 2013.

\bibitem{r_62}
Y.~Wu and Q.~Ji, ``Robust facial landmark detection under significant head
  poses and occlusion,'' in {\em Proceedings of the IEEE International
  Conference on Computer Vision}, pp.~3658--3666, 2015.

\bibitem{r_31}
X.~P. Burgosartizzu, P.~Perona, and P.~Doll{\'a}r, ``Robust face landmark
  estimation under occlusion,'' in {\em Computer Vision (ICCV), 2013 IEEE
  International Conference on}, pp.~1513--1520, 2013.

\bibitem{r_61}
S.~Tan, D.~Chen, C.~Guo, and et~al, ``A robust shape reconstruction method for
  facial feature point detection,'' {\em Computational intelligence and
  neuroscience}, vol.~2017, 2017.

\bibitem{r_47}
X.~Zhu and D.~Ramanan, ``Face detection, pose estimation, and landmark
  localization in the wild,'' in {\em Computer Vision and Pattern Recognition
  (CVPR), 2012 IEEE Conference on}, pp.~2879--2886, 2012.

\bibitem{esmaeilpour2019robust}
M.~Esmaeilpour, P.~Cardinal, and A.~L. Koerich, ``A robust approach for
  securing audio classification against adversarial attacks,'' {\em arXiv
  preprint arXiv:1904.10990}, 2019.

\bibitem{wang2016iterative}
Y.~Wang, W.~Zhang, L.~Wu, X.~Lin, M.~Fang, and S.~Pan, ``Iterative views
  agreement: An iterative low-rank based structured optimization method to
  multi-view spectral clustering,'' in {\em International Joint Conference on
  Artificial Intelligence (IJCAI)}, pp.~2153--2159, 2016.

\bibitem{wang2018multiview}
Y.~Wang, L.~Wu, X.~Lin, and J.~Gao, ``Multiview spectral clustering via
  structured low-rank matrix factorization,'' {\em IEEE transactions on neural
  networks and learning systems}, vol.~29, no.~10, pp.~4833--4843, 2018.

\end{thebibliography}
\vspace{12pt}
% \color{red}
% IEEE conference templates contain guidance text for composing and formatting conference papers. Please ensure that all template text is removed from your conference paper prior to submission to the conference. Failure to remove the template text from your paper may result in your paper not being published.

\end{document}